\documentclass[conference, letterpaper]{IEEEtran}

\IEEEoverridecommandlockouts
\usepackage{cite}
\usepackage{amsmath,amssymb,amsfonts}
\usepackage{algorithmic}
\usepackage{graphicx}
\usepackage{textcomp}
\usepackage{xcolor}
\usepackage{booktabs}
\usepackage{siunitx}
\usepackage[colorlinks=true, linkcolor=blue, citecolor=blue, urlcolor=blue]{hyperref}

\def\BibTeX{{\rm B\kern-.05em{\sc i\kern-.025em b}\kern-.08em
    T\kern-.1667em\lower.7ex\hbox{E}\kern-.125emX}}

\begin{document}

\title{Breaking the Average: Overcoming Convergent Defaults in Robotic Manipulation via Diagnostic Chain-of-Thought and Structured Guidance}

\author{\IEEEauthorblockN{Mohammed Daba}
\IEEEauthorblockA{\textit{School of Mechanical and Electrical Engineering} \\
\textit{University of Electronic Science and Technology of China}\\
Chengdu, Sichuan Province, P.R.of China \\
daba@std.uestc.edu.cn}
\and
\IEEEauthorblockN{Waleed Abdulrahman Saleh AlAkwa}
\IEEEauthorblockA{\textit{Department of Mechatronics Engineering} \\
\textit{University of Sana'a}\\
Sana'a, Yemen \\
waleed.alakwa@su.edu.ye}
}

\maketitle

\begin{abstract}
Recent Vision-Language-Action (VLA) models, such as $\pi_{0.5}$, have demonstrated impressive generalization capabilities by scaling behavior cloning across massive datasets. However, these models frequently suffer from \textit{distributional convergence}, a phenomenon where the policy regresses to ``safe'' but suboptimal mean behaviors due to mode averaging in the training data. This often results in generic grasp strategies, \textbf{Convergent Defaults}, that fail in constrained or long-tail scenarios. In this work, we propose a novel framework to identify and correct these behaviors without parameter updates. We introduce a three-stage methodology: (1) \textbf{Probing for Defaults}, identifying task-specific mode collapse; (2) \textbf{Diagnostic Root Cause Analysis}, utilizing the VLA's latent reasoning capabilities to verbalize the logic behind suboptimal actions; and (3) \textbf{Structured Guidance Injection}, where concrete, distinct alternatives are encoded into reusable ``Lens Prompts'' that shift the sampling distribution. We validate this approach using $\pi_{0.5}$ in ManiSkill3 simulation environments. We hypothesize that our method outperforms vanilla $\pi_{0.5}$ and standard Chain-of-Thought prompting by effectively breaking convergent defaults in complex manipulation tasks.
\end{abstract}


\section{Introduction}
The scaling of Foundation Models in robotics has led to the emergence of general-purpose Vision-Language-Action (VLA) models \cite{pi0_2024, rt2_2023}. Models such as $\pi_{0.5}$ utilize flow matching and massive datasets to learn versatile manipulation policies. However, a critical limitation remains: \textit{distributional convergence}.

\begin{figure}[t]
\centering
\includegraphics[width=\columnwidth]{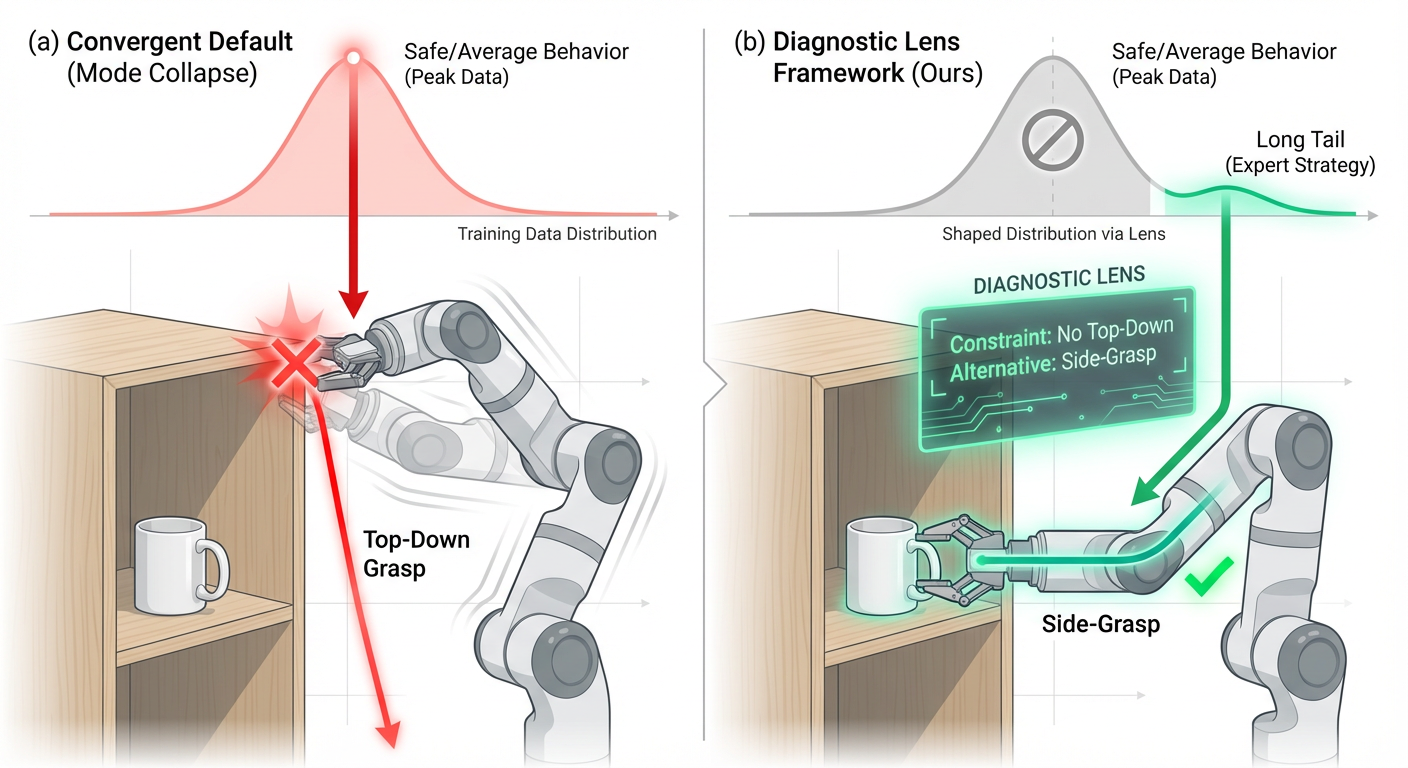}
\caption{\textbf{Breaking the Convergent Default.} VLA models often regress to the mean of their training data, causing failures in constrained environments. Our method identifies these modes and injects structured guidance (Lens) to shift the action distribution toward expert behaviors.}
\label{fig_teaser}
\end{figure}

In generative AI, this manifests as generic outputs where models produce statistically probable but safe responses to minimize loss \cite{wei2022chain}. In robotics, we observe an analogous phenomenon: \textbf{Mode-Averaged Action Priors}. When faced with ambiguity, VLA models often default to the kinematic mean of their training demonstrations. For example, a robot may approach an object from a generic top-down angle regardless of occlusion or fail to account for workspace constraints because the "average" training example assumes a clutter-free table. These ``Convergent Defaults'' are not hallucinations in the traditional sense; they are safe, high-likelihood predictions that fail in specific, context-heavy scenarios. We visually illustrate this phenomenon and our proposed solution in Fig. \ref{fig_teaser}.

Existing solutions largely focus on online self-correction \cite{li2024selfcorrecting} or reactive replanning \cite{sharma2022correcting}. While effective for error recovery, these methods are reactive. They do not address the underlying \textit{prior} that biased the robot toward the suboptimal action in the first place.

In this paper, we propose a framework to proactively reshape the VLA's action distribution. We formalize a methodology to:
1) \textbf{Isolate Convergent Defaults} by identifying where the model chooses a ``safe mean'' over a ``context-optimal'' action;
2) \textbf{Diagnose via Reasoning} by forcing the VLA to perform a ``Debug Mode'' analysis to verbalize latent assumptions; and
3) \textbf{Inject Structured Guidance} by constructing reusable prompt modules, termed \textbf{Lens Prompts}, that explicitly prohibit the default and provide concrete alternatives.

We apply this to $\pi_{0.5}$ in the ManiSkill3 \cite{maniskill2} simulation environment. Our contribution is a formalized pipeline for extracting expert-level performance from generalist VLAs without fine-tuning, shifting the paradigm from \textit{prompt engineering} to \textit{distribution shaping}.

% FIGURE 2: Top of Page 2
\begin{figure*}[t]
\centering
\includegraphics[width=\textwidth]{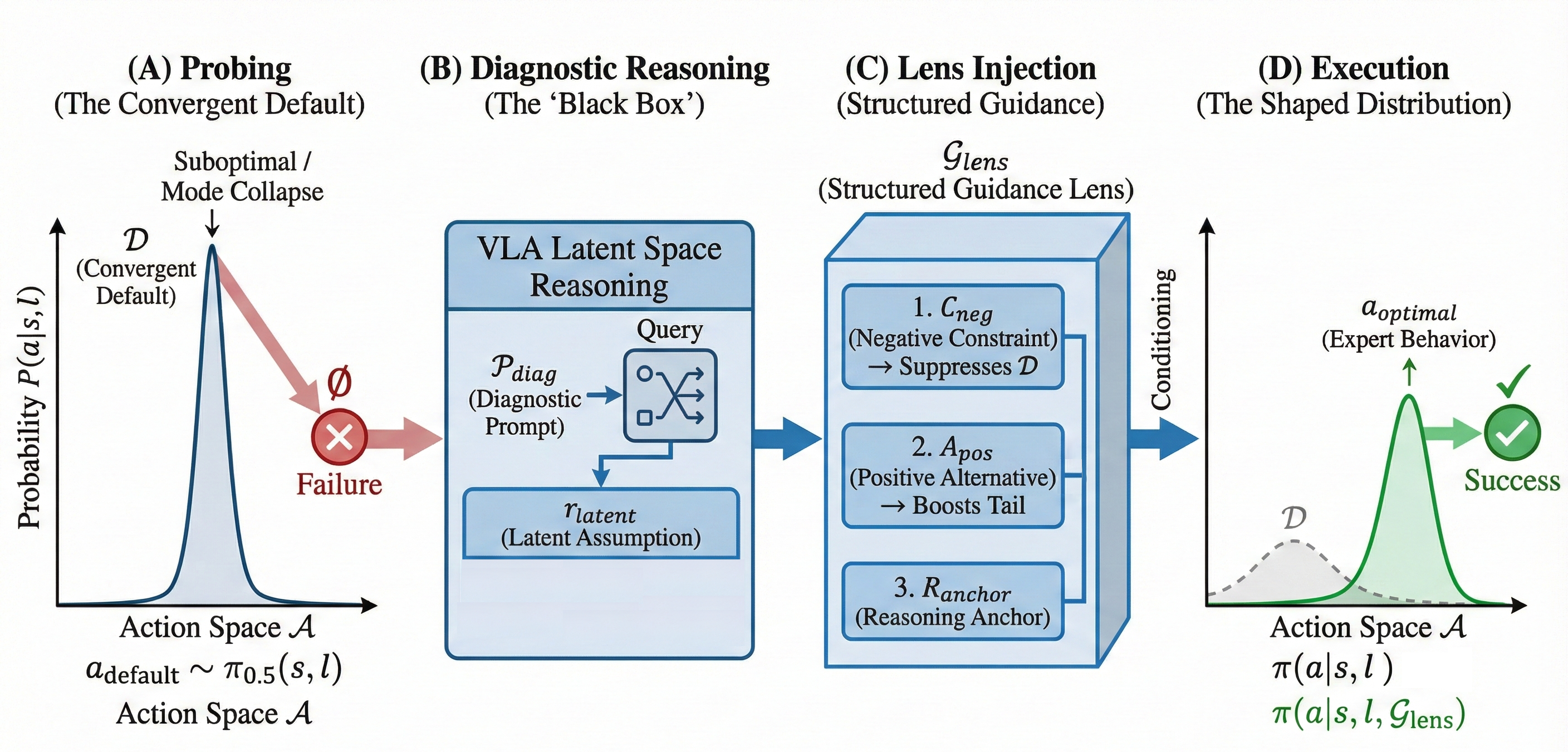}
\caption{\textbf{The Diagnostic Lens Framework.} We treat the VLA as a reasoning engine to diagnose its own action priors. By extracting the root cause of a default behavior, we construct a textual "Lens" that reshapes the flow matching distribution during inference.}
\label{fig_architecture}
\end{figure*}

\section{Related Work}

\subsection{Vision-Language-Action (VLA) Models}
Recent works like RT-2 \cite{rt2_2023} and $\pi_0$ \cite{pi0_2024} integrate visual perception with language reasoning. $\pi_{0.5}$, our baseline, utilizes flow matching to handle multi-modal distributions. Despite this, flow models can still suffer from mode collapse in the conditional generation step when prompts are underspecified \cite{ibc2022}. Our work addresses the gap where the model technically \textit{can} perform the task (if given the right steering), but the probability mass is centered on the wrong (generic) mode.

\subsection{Test-Time Adaptation \& Reasoning}
Methods like ``Inner Monologue'' \cite{innermonologue2022} and ``Self-Correcting VLAs'' \cite{li2024selfcorrecting} use LLMs to critique robot actions. However, these often rely on binary success/failure feedback. Our work differentiates itself by focusing on the \textbf{root cause analysis} of the prior distribution. We argue that generic feedback (``You failed, try again'') is less effective than \textbf{Structured Guidance}, a pre-computed set of positive constraints derived from analyzing the model's specific convergent defaults.

\section{Methodology}

Our goal is to modify the policy $\pi(a_t | s_t, l)$ (where $l$ is the language instruction) to a shaped policy $\pi(a_t | s_t, l, \mathcal{G}_{lens})$ without fine-tuning weights. We achieve this through the \textbf{Diagnostic Lens Framework}, a three-step cycle illustrated in Fig. \ref{fig_architecture}.

\subsection{Phase 1: Probing for Convergent Defaults}
We formalize a \textbf{Convergent Default} $\mathcal{D}$ as a high-density region in the action space where the model exhibits low variance (high confidence) despite suboptimal task performance. This occurs when the training distribution $P_{train}(a|s)$ is dominated by generic trajectories.

To identify $\mathcal{D}$, we probe the $\pi_{0.5}$ model with a zero-shot instruction $l_{base}$. We define the default behavior $a_{default}$ as the mode of the sampled actions. In constrained scenarios, $a_{default}$ often corresponds to the "kinematic average" of the training data (e.g., a straight-line approach) rather than a context-sensitive trajectory. Identification involves observing systematic failures where the physical constraints of the specific scene (e.g., obstacles, tight tolerances) contradict the statistical prior of the model.

\subsection{Phase 2: Diagnostic Root Cause Analysis}
Merely observing the failure is insufficient; we must identify the latent misconception driving the default. We exploit the VLA's capacity for verbalization to perform \textbf{Diagnostic Root Cause Analysis}. By freezing the visual context and prompting the model with a diagnostic query $\mathcal{P}_{diag}$, we extract the reasoning trace $r_{latent}$.

This process assumes that the VLA possesses the correct physical knowledge in its latent space but fails to activate it during standard action generation due to the strength of the convergent default. The diagnostic prompt forces the model to shift attention from \textit{action generation} to \textit{state explanation}. The output reveals whether the failure stems from a perceptual hallucination (e.g., "I did not see the shelf") or a heuristic bias (e.g., "I prioritized the top grasp because it is generally more stable"). This verbalized bias constitutes the \textbf{Root Cause}.

% FIGURE 3: Top of Page 3
\begin{figure*}[!t] 
\centering
\includegraphics[width=\textwidth]{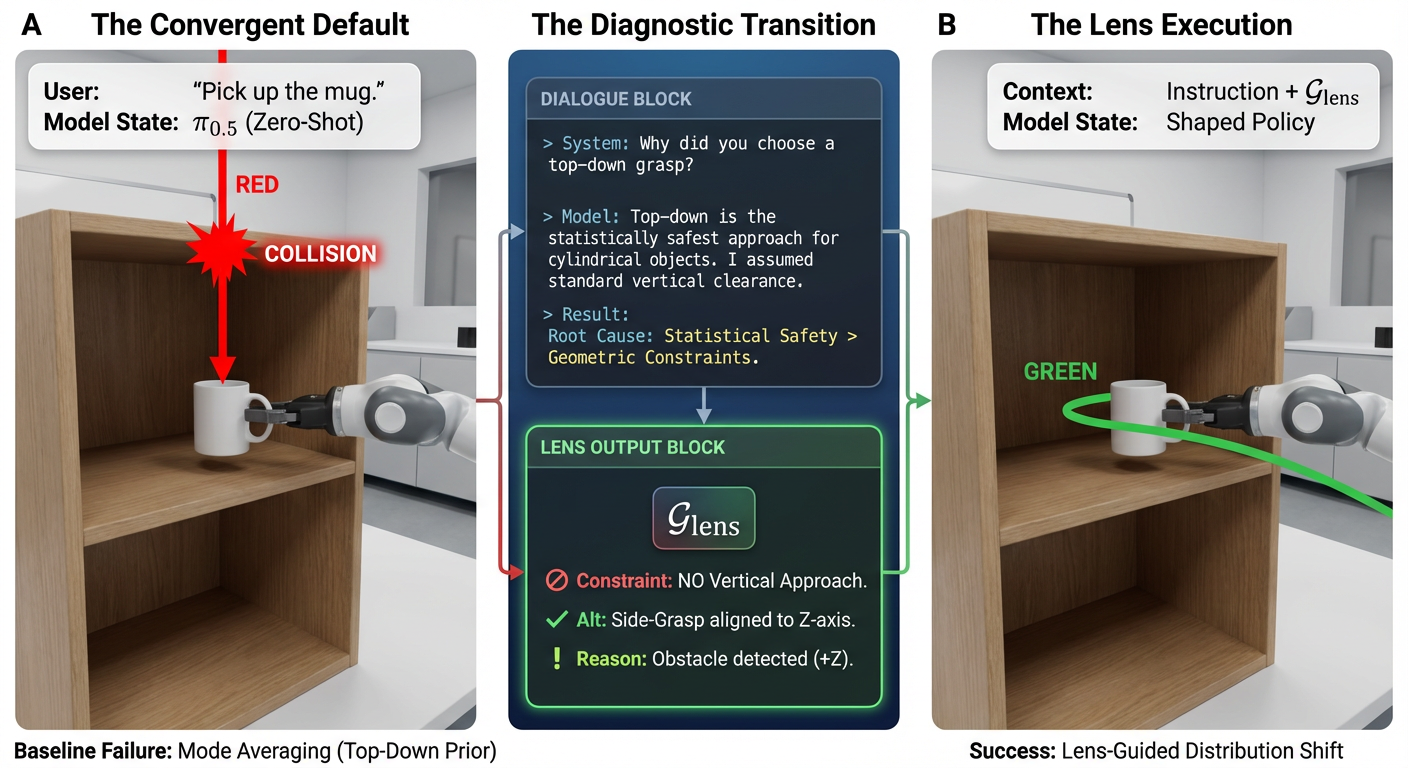}
\caption{\textbf{Illustrative Example: Constrained Mug Grasping.} This figure demonstrates the practical application of the methodology. The "Convergent Default" (top-down grasp) is identified, diagnosed as a statistical bias, and corrected via a specific Lens Prompt containing concrete alternatives.}
\label{fig_example}
\end{figure*}

\subsection{Phase 3: Structuring Guidance (The ``Lens'')}
The final phase involves synthesizing the root cause into a structured intervention, denoted as the \textbf{Lens} ($\mathcal{G}_{lens}$). The Lens is not merely a hint; it is a rigid context modifier designed to prune the default mode from the sampling distribution.

The Lens is constructed with three mandatory components:
\begin{enumerate}
    \item \textbf{Negative Constraint:} Explicitly forbidding the identified convergent default.
    \item \textbf{Positive Alternative:} Providing a concrete, geometrically distinct behavior to replace the default.
    \item \textbf{Reasoning Anchor:} grounding the alternative in the physical logic derived from Phase 2.
\end{enumerate}

Mathematically, this modifies the conditioning variable of the flow matching model, shifting the probability mass away from $\mathcal{D}$ and toward the long-tail region containing the optimal solution. An example of this construction is detailed in \textbf{Fig. \ref{fig_example}}.

\section{Experimental Setup}

\subsection{Simulation Environment}
We conduct experiments in \textbf{ManiSkill3} \cite{maniskill2}, a SAPIEN-based simulator known for high-fidelity physics and realistic rendering. This environment is crucial for testing VLA perception, as it supports diverse object geometries and physical interactions.

\subsection{Selected Benchmark Tasks}
We select tasks specifically designed to punish ``average'' behaviors (Convergent Defaults):
\begin{itemize}
    \item \textbf{Constrained Grasping:} Picking up objects from inside a shelf. The default behavior (top-down) causes collision with the shelf ceiling.
    \item \textbf{Precision Insertion:} Peg-in-hole with tight tolerances. The default behavior often lacks the necessary precision and re-orientation.
    \item \textbf{Articulated Cabinet Opening:} Opening a hinged cabinet door. The default behavior often attempts a linear pull (regression to the mean) rather than the required arced trajectory, causing the gripper to slip or the mechanism to jam.
\end{itemize}

\subsection{Baselines}
We compare our method against the following:
\begin{enumerate}
    \item \textbf{Vanilla $\pi_{0.5}$:} Zero-shot execution with standard instructions.
    \item \textbf{Generic Chain-of-Thought (CoT):} Prompting with ``Think step by step'' without domain-specific structured guidance.
    \item \textbf{Reflexion (Online):} A standard feedback loop where the model retries based on binary failure flags \cite{reflexion2023}.
\end{enumerate}

\subsection{Metrics}
\begin{itemize}
    \item \textbf{Success Rate (SR):} Percentage of successful task completions.
    \item \textbf{Reasoning Alignment:} A qualitative score measuring if the robot's generated text explanation matches its physical actions.
\end{itemize}

\section{Hypothesis and Expected Outcomes}

We hypothesize that \textbf{Vanilla $\pi_{0.5}$} will exhibit high failure rates in Constrained Grasping and Articulated tasks due to mode-averaging (convergent defaults). We expect \textbf{Generic CoT} to improve planning logic but potentially hallucinate affordances due to a lack of grounded constraints.

We anticipate that the \textbf{Diagnostic Lens Framework} will achieve the highest success rate. By explicitly shifting the probability mass toward the ``long-tail'' solutions (Concrete Alternatives) identified during the diagnostic phase, the model should effectively break out of the mode collapse. We expect to see a significant reduction in collisions for the constrained grasping task and improved trajectory adherence for articulated objects.

\section{Conclusion}
This work bridges the gap between generative distribution shaping and robotic control. By identifying and restructuring the ``Convergent Defaults'' of $\pi_{0.5}$, we enable the model to access expert-level behaviors hidden within its latent space. This approach offers a pathway to improve generalist robot performance in constrained environments without the high computational cost of parameter fine-tuning.

\bibliographystyle{IEEEtran}
\bibliography{references}

\end{document}